\documentclass[11pt]{article}

\usepackage[margin=1in]{geometry}
\usepackage{amsthm,amsmath,amsfonts,amssymb}
\usepackage[authoryear]{natbib}
\usepackage[colorlinks,citecolor=blue,urlcolor=blue,linkcolor=blue,bookmarksnumbered]{hyperref}
\usepackage{bookmark}
\usepackage{graphicx}
\usepackage{cleveref}

\usepackage{algorithm}
\usepackage{algorithmic}

\usepackage{setspace}
\usepackage{authblk}

\makeatletter
\let\old@paragraph\paragraph
\def\paragraph{\@ifstar\sada@paragraph@s\sada@paragraph@n}
\def\sada@paragraph@s#1{\old@paragraph*{#1.}}
\def\sada@paragraph@n#1{\old@paragraph{#1.}}
\makeatother

\theoremstyle{plain}

\theoremstyle{definition}

\usepackage{bm}
\usepackage{booktabs}
\usepackage{xcolor}
\usepackage{tcolorbox}
\tcbuselibrary{breakable}
\usepackage{multirow}
\usepackage{graphicx}
\usepackage{subcaption}

\def\0{\mathbf{0}}

\newcommand{\quotes}[1]{``#1''}

\def\boxit#1{\vbox{\hrule\hbox{\vrule\kern6pt\vbox{\kern6pt#1\kern6pt}\kern6pt\vrule}\hrule}}

\def\boxit#1{\vbox{\hrule\hbox{\vrule\kern6pt\vbox{\kern6pt#1\kern6pt}\kern6pt\vrule}\hrule}}

\def\boxit#1{\vbox{\hrule\hbox{\vrule\kern6pt\vbox{\kern6pt#1\kern6pt}\kern6pt\vrule}\hrule}}

\newcommand{\methodname}{Frontier Learning}

\title{Learning the Pareto Frontier of Predictive Models under Distribution Shift}

\author[1]{Yiming Dong}

\author[1,2]{Jiwei Zhao}

\author[3]{Yang Lu}

\affil[1]{Department of Biostatistics \& Medical Informatics, University of Wisconsin--Madison}

\affil[2]{Department of Statistics, University of Wisconsin--Madison}

\affil[3]{Department of Biomedical Engineering, University of Wisconsin--Madison}

\date{}

\begin{document}

\maketitle

\begin{abstract}
\noindent
Modern machine learning pipelines increasingly rely on reusing pretrained and foundation models across downstream tasks. 
These pretrained models can differ not only in performance but also in how they can be used: some only provide black-box predictions, while others may permit white-box access to internal representations that can be probed or fine-tuned. 
When it is deployed to the target in the presence of distribution shift, no single strategy, including but not limited to, zero-shot application, fine-tuning, directly training a target-specific model, is uniformly the best, and the right choice usually depends on the source-target shift that is typically unknown in advance. 
In this work, we propose Frontier Learning, a framework that treats a library of candidate models spanning different training histories and access regimes as complementary sources of information rather than mutually exclusive alternatives. 
Frontier Learning constructs a unified target-domain feature by concatenating internal representations from white-box candidates as well as prediction outputs from black-box candidates, then fits a lightweight, regularized supervised learner on this concatenated representation using labeled target data. 
Because the resulting hypothesis class contains predictors obtained by zero-shot reuse, fine-tuning, and direct training as special cases, empirical risk minimization over the frontier learner is guaranteed to be no worse, on the training sample, than any individual baseline.
We evaluate the framework in simulations spanning varying degrees of source-target compatibility and in two real-world distribution-shift settings: visual domain adaptation on DomainNet/VisDA and clinical mortality prediction across intensive care unit domains using MIMIC-IV-Notes. 
Across all settings, Frontier Learning matches or outperforms the strongest individual reuse strategy, with the largest gains arising precisely when no single baseline is reliable across the range of shift considered. 
These results suggest that, in an era of proliferating black-box and white-box foundation models, learning to combine candidate representations and predictions is a practical, statistically grounded alternative to committing to a single pretrained model for reuse.
\end{abstract}


\newpage
\medskip
\section{Introduction}
Machine learning is shifting from the classical paradigm of training one task-specific predictor on one dataset toward a new paradigm of reusing and adapting large pretrained models across many downstream tasks \citep{bommasani:opportunities}.
This shift is especially visible in generative AI and large language models (LLMs), which can perform new tasks through prompting or in-context examples without task-specific gradient updates
\citep{brown:language,openai:gpt4}.
In practice, however, pretrained models differ not only in predictive performance but also in accessibility: some are black-box systems that return predictions through an API, whereas others are white-box or open-weight systems whose representations and parameters can be inspected or modified \citep{sun:black-box}.
For downstream biomedical tasks, the goal is no longer only to fit a model on the target data, but also to decide how to use external models that may have already learned from large-scale data but whose relevance to the target population is uncertain \citep{bommasani:holistic}.

These disparate access regimes lead naturally to three common predictive strategies. 
The first is zero-shot reuse, in which a pretrained or externally hosted model is applied directly to the target population without task-specific retraining \citep{brown:language,openai:gpt4}.
This strategy is attractive when internal model access is unavailable or retraining is computationally prohibitive, but its predictions can be biased or miscalibrated when the target distribution differs from the model's training or alignment distribution \citep{guo:calibration}.
The second strategy is fine-tuning, in which a white-box model is adapted to the target task by updating all parameters or a parameter-efficient subset, such as adapters or low-rank updates \citep{houlsby:parameter,hu:lora}.
Fine-tuning can effectively exploit pretrained representations, but it requires choosing which model to adapt and can overfit when labeled target data are limited \citep{pan:survey}.
The third strategy is direct training using only target-domain data.
This strategy avoids reliance on a potentially misleading source model but discards pretrained knowledge and can have high variance in small target samples \citep{quinonero:dataset}.

None of these strategies is uniformly preferable across all scenarios. 
Modern model ecosystems contain many strong but heterogeneous candidates, including GPT \citep{openai:gpt4}, Llama \citep{grattafiori:llama}, Qwen \citep{bai:qwen}, DeepSeek \citep{liu:deepseek-v3}, and domain-specific models \citep{singhal:large}.
These models can differ in pretraining data, architecture, post-training procedure, calibration, reasoning behavior, access mode, and target-domain reliability \citep{bommasani:holistic}.
The difficulty is amplified under distribution shift, where a model developed in one population, cohort, hospital, imaging protocol, or documentation system may be deployed in another setting with different covariates, outcome prevalence, measurement processes, annotation practices, or selection mechanisms \citep{quinonero:dataset}.
Thus, the analyst faces a dilemma: a black-box model may be convenient but unreliable, a fine-tuned white-box model may be powerful but over-adapted, and a directly trained model may be locally aligned but data inefficient.

We propose Frontier Learning, a framework for learning over a frontier of predictive models under distribution shift.
Rather than treating zero-shot, fine-tuning, and direct models as mutually exclusive alternatives, Frontier Learning treats them as candidate sources of information whose usefulness can be learned from target-domain labels. The framework accommodates both white-box and black-box candidate models.
We use the term \quotes{frontier} to emphasize that these candidate models represent different tradeoffs among target fit, source reliance, labeled-sample efficiency, computational cost, and access to model internals.
The goal is not to decide which model is the best, but to use the available target labels to learn how the candidate models should contribute to the final predictor.
Operationally, Frontier Learning begins with a library of candidate models generated under different training and access regimes.
For white-box candidates, the framework can use internal representations, such as bottleneck or penultimate-layer features. 
For black-box candidates, the framework uses prediction outputs. 
A lightweight supervised learner is trained on the resulting collection of representations and predictions using labeled target-domain data. 
This access-adaptive design allows Frontier Learning to synthesize information from zero-shot, fine-tuned, and directly trained models, together with prediction outputs from black-box models, within a single predictive framework.

The main statistical advantage of Frontier Learning is that standard predictive strategies remain recoverable as special cases of the learned frontier predictor.
If the target learner uses only a source-trained candidate, it recovers a zero-shot-style predictor; 
if it uses only a fine-tuned candidate, it recovers a fine-tuning-based predictor; 
and if it uses only a directly trained candidate, it recovers direct target training. 
Consequently, the frontier hypothesis class contains these baseline classes as special cases, yielding a no-worse empirical-risk property on the labeled target sample used to train the frontier learner. 
This property should not be interpreted as an unconditional guarantee of lower test risk. 
Rather, it provides a structural basis for cautious model composition, with regularization and validation used to control finite-sample generalization.

This paper makes four contributions. 
First, we formulate prediction under distribution shift as a model-frontier learning problem motivated by the modern foundation-model ecosystem, where candidate models may differ in both training regime (zero-shot, fine-tuned, or directly trained) and access regime (black-box or white-box).
Second, we introduce a unified framework that combines internal representations from accessible models with prediction outputs from inaccessible models.
Third, we establish a no-worse empirical-risk property through hypothesis-class containment, showing that standard model-construction strategies are recoverable as special cases.
Fourth, we evaluate the proposed method in simulations and real-data studies, including visual domain shift and clinical note-based mortality prediction across intensive-care-unit domains, demonstrating that learning over the model frontier can match or improve upon strong individual candidates while providing a practical route to more cautious model reuse under distribution shift.
In the era of LLMs and foundation models, Frontier Learning offers a statistical route from choosing a single pretrained model to learning over a heterogeneous frontier of candidates, enabling more reliable reuse of large-scale AI models under distribution shift.

\section{Related works}
\methodname\ is related to transfer learning and domain adaptation, but it differs by learning over a frontier of candidate models rather than adapting a single source model.
Transfer learning studies how knowledge from a source domain can improve performance in a target domain when target data are limited \citep{pan:survey}.
Dataset shift and domain-adaptation theory show that source-domain performance alone does not guarantee target-domain performance when the source and target distributions differ \citep{quinonero:dataset}.
Deep domain-adaptation methods often address this problem by learning domain-invariant or target-adapted representations, as in domain-adversarial neural networks \citep{ganin:domain-adversarial}.
Source-free methods relax the need for source data by adapting target features while preserving a source-trained hypothesis \citep{liang:we}.
Multi-source domain adaptation further considers how to use multiple source domains, often through feature alignment or distribution matching across domains \citep{peng:moment}.
Frontier Learning differs from these approaches because it does not aim to produce a single adapted representation; instead, it uses labeled target data to learn how to combine zero-shot, fine-tuned, directly trained, and black-box candidate models within one predictive frontier.

Frontier Learning is related to prediction-side calibration and prediction-powered inference because all three treat model outputs as useful but imperfect information.
Modern neural networks are often poorly calibrated, and post-hoc methods such as temperature scaling can improve the agreement between predicted probabilities and observed frequencies without changing the underlying predictor \citep{guo:calibration}.
Under domain shift, calibration becomes more difficult because source-target mismatch affects both predictive accuracy and uncertainty estimation; methods such as TransCal therefore recalibrate transferred models to improve reliability under adaptation \citep{wang:transferable}.
Prediction-powered inference takes a broader statistical view, using machine predictions together with a smaller set of gold-standard labels to obtain valid inference for population quantities without assuming that the predictive model is correct \citep{angelopoulos:prediction}.
Frontier Learning shares this cautious view of machine predictions, but pursues a different goal. 
Rather than recalibrating a fixed predictor, Frontier Learning treats predictions and representations from multiple candidate models as supervised target-domain features, allowing the final learner to use and combine transferred information according to target labels.

Frontier Learning is statistically closest to ensemble learning, stacking, and Super Learner, but the meanings of the model library are different. Stacked generalization and stacked regression learn a second-stage model that combines predictions from multiple base learners \citep{wolpert:stacked}.
The Super Learner framework provides a loss-based ensemble procedure with oracle-type guarantees relative to a prespecified library of candidate learners \citep{van:super}.
Frontier Learning inherits this statistical spirit, but its library is not merely a set of algorithms trained on the same target dataset.
Instead, the library consists of models produced under different training histories, source-target relationships, and access regimes, including zero-shot models, fine-tuned models, directly trained models, and black-box models.
When internal representations are available, Frontier Learning also moves beyond ordinary prediction stacking by combining representation-level information that may be discarded by final prediction heads.

Lastly, Frontier Learning also connects to Pareto and multi-objective perspectives because its candidate models encode different practical tradeoffs. 
In multi-objective optimization, a Pareto frontier represents solutions for which no objective can be improved without worsening another objective \citep{deb:fast}. 
In multi-task and multi-objective learning, this perspective has been used to reason about competing performance criteria and tradeoff surfaces rather than a single scalar optimum \citep{sener:multi-task}. 
\methodname\ uses the term ``frontier'' in a more model-centric sense: zero-shot reuse, fine-tuning, and direct training represent different tradeoffs among target fit, source reliance, sample efficiency, computation, and model access. 
The method then uses labeled target data to learn over this induced predictive frontier, recovering strong individual candidates when appropriate and exploiting complementary candidates when possible.

\section{Method}

\subsection{Problem setup}
Let $\mathcal{P}_T$ denote the target distribution over input--outcome pairs $(X,Y)\in \mathcal X\times \mathcal Y$. 
We observe a labeled target sample $\mathcal D_T=\{(x_i,y_i)\}_{i=1}^n$ drawn from $\mathcal{P}_T$. 
The goal is to construct a target-domain predictor $f:\mathcal X\rightarrow \mathcal A$, where $\mathcal A$ is the prediction space. 
For regression, $\mathcal A=\mathbb R$;  
for binary classification, $\mathcal A=[0,1]$; 
and for $K$-class classification, $\mathcal A=\Delta^{K-1}$, the probability simplex over $K$ classes. 
Given a task-specific loss function $\ell$, the target risk is
\[
R_T(f)=\mathbb E_{(X,Y)\sim \mathcal P_T}\{\ell(Y,f(X))\}.
\]

The source and target domains exhibit a distribution shift while still sharing transferable predictive information (Fig.~\ref{fig:FLillusion}a). The objective is to effectively leverage this transferable information while remaining robust to distribution shift using the available labeled target data.

In addition to the labeled target sample, as illustrated in Fig.~\ref{fig:FLillusion}b, we assume the access to a library of
candidate models that may be obtained under different model construction regimes.
Some candidates are source-trained models applied directly to the target population without any target-domain adaptation, which we refer to as zero-shot models. 
Some candidates are initialized from pretrained source checkpoints and adapted using target-domain labels, which we refer to as fine-tuned models. 
Others are trained using only target-domain data, which we refer to as directly trained models. 
The library may also contain models trained on different source domains, models with different architectures, or proprietary predictors accessible only through queries.

The central difficulty is that no candidate model is uniformly the best under distribution shift. 
A zero-shot model may encode useful source-domain structure but could be miscalibrated in the target population. 
A fine-tuned model may adapt source representations but can overfit when labeled target data are limited. 
A directly trained model is directly aligned with target labels but may have high variance. 
\methodname\ treats these candidate models as points on a common predictive frontier and uses labeled target data to learn how their available information should be combined.

\subsection{Candidate models and access regimes}

\begin{figure}[ht]
    \centering
    \includegraphics[width=1.0\textwidth]{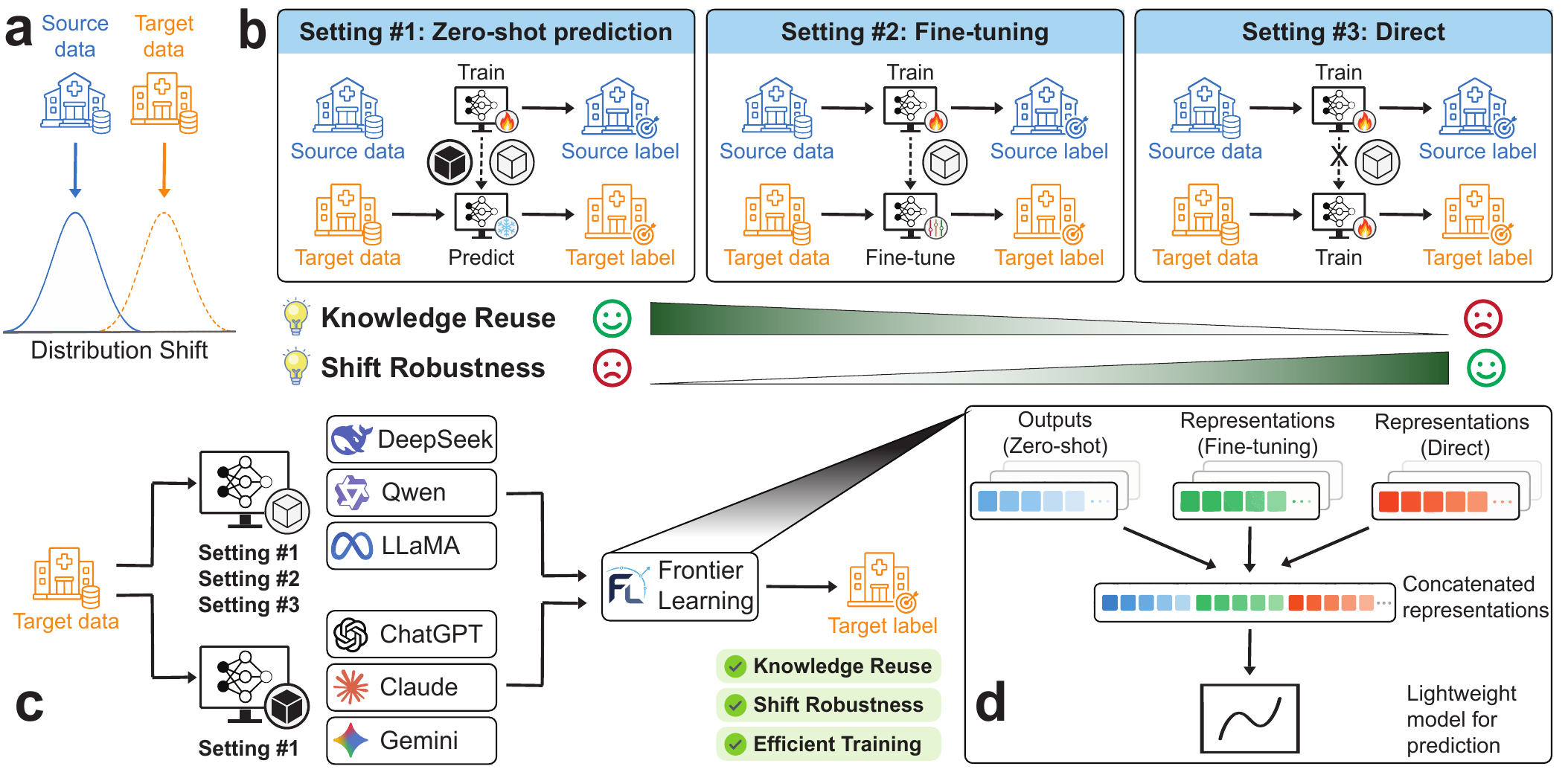}
    \caption{
    \textbf{Overview of the Frontier Learning framework for robust prediction under distribution shift.}
    \textbf{(a)}
    Source and target domains exhibit distribution shift while sharing transferable predictive information.
    \textbf{(b)}
    Three practical transfer learning settings are considered: zero-shot prediction, fine-tuning, and direct, illustrating the trade-off between knowledge reuse and robustness to distribution shift.
    \textbf{(c)} 
    \methodname\ unifies predictions from multiple foundation models trained under different transfer settings to leverage their complementary strengths.
    \textbf{(d)}
    Rather than combining model outputs directly,  \methodname\ concatenates latent representations from white-box models and prediction outputs from black-box models, and trains a lightweight predictor on the resulting frontier representation using labeled target data.
    }
    \label{fig:FLillusion}
\end{figure}

As illustrated in Fig.~\ref{fig:FLillusion}c, candidate models differ not only in how they are constructed, but also in what information they expose. We distinguish between white-box and black-box models.
For a white-box model $f_m$, we assume access to an internal representation. 
For neural-network models, this representation may be a bottleneck feature, a pooled embedding, or a penultimate-layer activation. 
We write
\[
f_m(x)=c_m(r_m(x)),
\]
where $r_m(x)\in \mathbb R^{d_m}$ is the extracted latent representation with $d_m$ as the representation dimensionality, and $c_m(\cdot)$ is the final prediction of the model. 
\methodname\ uses latent representations $r_m(x)$ because they preserve task-relevant information about the input that is not fully captured by the final prediction output.
For a black-box candidate model, internal representations and parameters are not available. 
In this case, \methodname\ uses the model's prediction output, denoted by $q_m(x)\in \mathbb R^{a_m}$.
For regression, $q_m(x)$ is typically a scalar prediction. 
For binary classification, it may be a logit, probability, or risk score. 
For multi-class classification, it may be a vector of logits or predicted class probabilities.
When logits are available, we use them in preference to normalized probabilities because they preserve pre-softmax score information.
If only hard labels are returned, they can be encoded as categorical indicators, although this discards uncertainty information contained in continuous scores.
For black-box models, this limitation can be partially mitigated by querying the model multiple times under non-deterministic decoding and using the empirical frequency of returned labels as a soft approximation to the model's predictive distribution, following the sampling-and-aggregation principle used in black-box LLM uncertainty estimation \citep{lin:generating}.
Thus, we define the signal contributed by candidate \(m\) as:
\[
\psi_m(x)=
\begin{cases}
r_m(x), & \text{if candidate } m \text{ is used as a white-box model},\\[3pt]
q_m(x), & \text{if candidate } m \text{ is used as a black-box model}.
\end{cases}
\]
In the experiments in this paper, we use hidden representations for white-box models and prediction outputs for black-box models unless otherwise specified.
As illustrated in Fig.~\ref{fig:FLillusion}d, \methodname\ constructs a unified target-domain representation by concatenating the signals from all candidate models:
\[
\Psi(x)=\left[\psi_1(x);\psi_2(x);\cdots;\psi_M(x)\right]\in \mathbb R^D,
\]
where
\[
D=\sum_{m=1}^M \dim\{\psi_m(x)\}.
\]
We refer to \(\Psi(x)\) as the \emph{frontier representation}. 
It contains the information made available by the candidate models, including hidden representations from white-box models and prediction outputs from black-box models.
All candidate models are held fixed during the \methodname\ stage.

\subsection{Target-supervised frontier learner}
Given the frontier representation \(\Psi(x)\), \methodname\ trains a lightweight supervised learner
\[
g_\theta:\mathbb R^D\rightarrow \mathcal A
\]
using labeled target-domain data. The final Frontier Learning predictor is
\[
f_{\mathrm{FL}}(x)=g_{\widehat\theta}(\Psi(x)),
\]
where the parameter \(\theta\) is estimated by regularized empirical risk
minimization:
\[
\widehat\theta
=
\arg\min_{\theta\in\Theta}
\left[
\frac{1}{|\mathcal I_{\mathrm{tr}}|}
\sum_{i\in\mathcal I_{\mathrm{tr}}}
\ell\{y_i,g_\theta(\Psi(x_i))\}
+
\lambda \Omega(\theta)
\right],
\]
where \(\mathcal I_{\mathrm{tr}}\) indexes the target observations used to train
the frontier learner, \(\Omega(\theta)\) is a regularization penalty, and
\(\lambda\geq 0\) is selected using validation data or cross-validation.

For regression tasks, we use a regularized linear frontier learner,
\[
g_\theta(z)=\beta_0+\beta^\top z,
\]
with coefficients estimated by
\[
(\widehat\beta_0,\widehat\beta)
=
\arg\min_{\beta_0,\beta}
\left[
\frac{1}{|\mathcal I_{\mathrm{tr}}|}
\sum_{i\in\mathcal I_{\mathrm{tr}}}
\{y_i-\beta_0-\beta^\top \Psi(x_i)\}^2
+
\lambda \|\beta\|_2^2
\right].
\]
For binary classification tasks, we use a regularized logistic frontier learner,
\[
\mathbb P(Y=1\mid X=x)
=
\sigma\{\beta_0+\beta^\top \Psi(x)\},
\]
where \(\sigma(t)=1/(1+\exp(-t))\) and the parameters are estimated by penalized binary cross-entropy.
For multi-class classification tasks, we use a regularized multinomial logistic frontier learner,
\[
\mathbb P(Y=k\mid X=x)
=
\frac{
\exp(w_k^\top \Psi(x)+b_k)
}{
\sum_{\ell=1}^K \exp(w_\ell^\top \Psi(x)+b_\ell)
},
\qquad k=1,\ldots,K,
\]
with parameters estimated by minimizing the cross-entropy loss:
\[
(\widehat W,\widehat b)
=
\arg\min_{W,b}
\left[
\frac{1}{|\mathcal I_{\mathrm{tr}}|}
\sum_{i\in\mathcal I_{\mathrm{tr}}}
\ell_{\mathrm{CE}}\{y_i,\mathrm{softmax}(W\Psi(x_i)+b)\}
+
\lambda \|W\|_F^2
\right].
\]

We deliberately use shallow regularized learners. 
The goal of \methodname\ is not to train a new high-capacity model on top of all candidates, but to learn a stable target-domain combination of candidate information.
Regularization encourages the frontier learner to downweight candidate signals that are uninformative or harmful under target-domain shift.
Because this stage only fits a lightweight shallow learner on fixed candidate representations or predictions, \methodname\ adds little computational overhead relative to training or fine-tuning the candidate models themselves.

\subsection{\methodname\ as a unified framework}
The construction above makes \methodname\ a unified framework rather than a single transfer-learning strategy. 
Standard approaches such as zero-shot reuse, fine-tuning, and direct training correspond to using different candidate models.
\methodname\ instead places these models in a shared frontier representation and lets the target-domain learner determine how much each candidate model should contribute.
To see this, we partition the candidate models into white-box candidates $\mathcal{W}$ and black-box candidates $\mathcal{B}$.
For $m\in\mathcal{W}$, the candidate contributes an internal representation $r_m(x)$;
for $m\in\mathcal B$, it contributes a prediction output $q_m(x)$. 
The frontier representation can therefore be written as:
\[
\Psi(x)
=
\left[
\{r_m(x):m\in\mathcal W\};
\{q_m(x):m\in\mathcal B\}
\right].
\]
The final learner $g_\theta$ is then trained on this representation using labeled target-domain data. 
Thus, \methodname\ covers both representation-level learning from white-box models and prediction-level learning from black-box models.

This formulation contains several familiar procedures as limiting cases. 
If all candidate models are black-box and only prediction outputs are used, then \methodname\ reduces to supervised prediction-level stacking. 
If all candidate models are white-box and hidden representations are used, then \methodname\ becomes a representation-level learner over the concatenated feature spaces of
the candidate models. 
If the final learner uses only one candidate block, then the resulting predictor reduces to a single-model strategy: using only a zero-shot block recovers a zero-shot-style predictor, using only a fine-tuned block recovers a fine-tuning-based predictor, and using only a directly trained block recovers direct training.
The precise meaning of \quotes{recovers} depends on what information from each candidate is included in $\psi_m(x)$. 
If the prediction output $q_m(x)=f_m(x)$ is included, then the original candidate prediction is available directly to the frontier learner. 
If only the internal representation $r_m(x)$ is included, exact recovery of the original candidate predictor $f_m(x)=c_m(r_m(x))$ requires that the candidate head $c_m(\cdot)$ belongs to the class of frontier learners used in the final stage. 
Methodology-wise, \methodname\ does not require the analyst to decide in advance whether zero-shot reuse, fine-tuning, or direct training will be most effective.
Instead, it embeds these strategies as candidate components within a larger target-supervised predictor.

In theory, because of its architecture, \methodname\ is guaranteed to achieve the training risk no worse than either of its candidate models.
To be more formal, denote $f_{\mathrm{zs}}(x)$ the zero-shot model output (internal representation if white-box or prediction output if black-box), and, similarly, $f_{\mathrm{ft}}(x)$ the fine-tuning model output and $f_{\mathrm{dir}}(x)$ the output of the directly trained model.
For simplicity, denote $f_{\mathrm{baseline}}(x)$ the output of a ``baseline'' candidate model where the $\rm{baseline}$ could be either $\rm{zs}$, $\rm{ft}$, $\rm{dir}$ or any other candidate model being considered.
In addition, we also denote $\mathcal{F}_{\rm{baseline}}$ the function class of a particular baseline candidate model and $\mathcal{F}_{\rm{FL}}$ the function class of the \methodname. Since \methodname\ hypothesis class contains all baseline model classes as special cases,
\[
\mathcal F_{\mathrm{baseline}}
\subseteq
\mathcal F_{\mathrm{FL}},
\]
empirical risk minimization over the \methodname\ class satisfies
\[
\inf_{f\in\mathcal F_{\mathrm{FL}}}
\widehat R_T(f)
\le
\inf_{f\in\mathcal F_{\mathrm{baseline}}}
\widehat R_T(f).
\]
Thus, \methodname\ is guaranteed to achieve training risk no worse than any individual baseline candidate model, including either $\rm{zs}$, $\rm{ft}$ or $\rm{dir}$.

\subsection{Illustrative linear case}
We give a simple linear example to clarify how \methodname\ combines candidate information. 
Consider the linear-predictor scale used in linear regression, logistic regression, or multinomial logistic regression. 
Suppose that the frontier learner is linear in the candidate signals. 
Then the linear predictor can be written as: 
\[
\eta_{\mathrm{FL}}(x)
=
\beta_0
+
\sum_{m\in\mathcal W}\alpha_m^\top r_m(x)
+
\sum_{m\in\mathcal B}\gamma_m^\top q_m(x),
\]
where \(\alpha_m\) weights the representation block from white-box model \(m\), and \(\gamma_m\) weights the prediction-output block from black-box model \(m\). 
For regression, the final prediction is \(f_{\mathrm{FL}}(x)=\eta_{\mathrm{FL}}(x)\). 
For binary classification, \(\eta_{\mathrm{FL}}(x)\) is passed through a logistic link. 
For multi-class classification, the same construction is applied class-wise to define the multinomial logits.

This expression shows how candidate models enter the frontier learner. 
If \(\alpha_m=0\), the representation from white-box model \(m\) is ignored. 
If \(\gamma_m=0\), the prediction output from black-box model \(m\) is ignored. 
If only one block has nonzero weight, the frontier learner reduces to a single-candidate predictor. 
If multiple blocks have nonzero weights, the final predictor combines complementary information across candidates.

For further intuition, suppose that each white-box representation is an affine transformation of a candidate-specific backbone representation:
\[
r_m(x)=A_m\phi_m(x)+a_m,
\qquad m\in\mathcal W,
\]
where \(\phi_m(x)\) denotes the backbone feature representation of model \(m\). 
Substituting this expression into the linear predictor gives: 
\[
\eta_{\mathrm{FL}}(x)
=
\beta_0
+
\sum_{m\in\mathcal W}\alpha_m^\top
\{A_m\phi_m(x)+a_m\}
+
\sum_{m\in\mathcal B}\gamma_m^\top q_m(x).
\]
Equivalently,
\[
\eta_{\mathrm{FL}}(x)
=
\widetilde\beta_0
+
\sum_{m\in\mathcal W}\widetilde\alpha_m^\top \phi_m(x)
+
\sum_{m\in\mathcal B}\gamma_m^\top q_m(x),
\]
where
\[
\widetilde\alpha_m=A_m^\top\alpha_m,
\qquad
\widetilde\beta_0
=
\beta_0+\sum_{m\in\mathcal W}\alpha_m^\top a_m.
\]
Thus, in the linear case, \methodname\ can be interpreted as learning a target-domain linear predictor over the union of candidate backbone feature
spaces, together with any available black-box prediction outputs.
This example also clarifies why representation-level \methodname\ is more general than ordinary prediction stacking. 
Prediction stacking can only weight the final outputs \(q_m(x)\). 
In contrast, representation-level \methodname\ can use intermediate features \(r_m(x)\) or \(\phi_m(x)\), which may contain task-relevant information that was not expressed in the model's final prediction. 
The mixed-access version combines both views: predictions are used when models are black-box, and internal representations are used when models are white-box.

\section{Simulation Studies}

\subsection{Overview}

We first evaluate \methodname\ in a controlled regression experiment designed to study how the method behaves as the alignment between source and target information varies. The purpose of this simulation is not to mimic a specific application, but to isolate a central feature of the proposed framework: zero-shot reuse, fine-tuning, and direct training may each be useful in different regimes, and a target-supervised frontier learner should be able to combine their complementary information rather than selecting one strategy in advance.

We consider one source domain and one target domain. For the source domain, we generate \(n_S=10{,}000\) samples and split them into training, validation, and test sets of sizes \(8000\), \(1000\), and \(1000\), respectively. For the target domain, we generate \(n_T=5000\) samples and split them into training, validation, and test sets of sizes \(3000\), \(1000\), and \(1000\), respectively. All simulation results are averaged over 100 independent repetitions.

All candidate models use the same two-layer multilayer perceptron architecture. For an input vector \(x\in\mathbb R^{30}\), the first hidden layer is \(h_1(x)=\sigma(W_1x+b_1)\), the second hidden layer is \(h_2(x)=\sigma(W_2h_1(x)+b_2)\), and the final prediction is \(f(x)=W_3h_2(x)+b_3\), where \(\sigma(t)=\max(t,0)\) is the ReLU activation function. Both hidden layers have dimension 128. The second hidden layer \(h_2(x)\) is used as the hidden representation for white-box candidates.

Models are trained using mean squared error loss. We use AdamW with initial learning rate \(10^{-3}\), batch size 1024, and weight decay \(10^{-4}\), together with a cosine learning-rate decay schedule. For fine-tuning, we use a smaller learning rate of \(10^{-4}\). All models are trained for at most 200 epochs with early stopping based on validation MSE and patience 30.

We compare \methodname\ against three standard strategies. The first is source zero-shot, where the source-trained model is applied directly to the target domain without adaptation; this candidate is treated as black-box, so only its scalar prediction output is used. The second is fine-tuning, where the model is initialized from the source checkpoint and fully fine-tuned on the target training set; this candidate is treated as white-box, so its hidden representation is used. The third is direct target training, where a model is trained from random initialization using only target-domain data; this candidate is also treated as white-box. For \methodname, we construct the frontier representation as \(\Psi(x)=[q_{\mathrm{zs}}(x);r_{\mathrm{ft}}(x);r_{\mathrm{dir}}(x)]\), where \(q_{\mathrm{zs}}(x)\) is the scalar prediction from the source zero-shot model, and \(r_{\mathrm{ft}}(x)\) and \(r_{\mathrm{dir}}(x)\) are 128-dimensional hidden representations from the fine-tuned and directly trained models. Thus, the frontier representation has dimension \(1+128+128=257\). A regularized linear regression model is then trained on the target training set, with hyperparameters selected using the target validation set.

Because the task is regression, we evaluate all methods using MSE, RMSE, MAE, and \(R^2\). We emphasize MSE in the main text and Figure~\ref{fig:simulation}, and report the full set of metrics in Table~\ref{tab:simulation_full}.

\subsection{Simulation setup}

For both domains, the covariate dimension is \(d=30\). Source and target covariates are independently sampled from the same Gaussian distribution \(X\sim N(\mu,\Sigma)\), where \(\mu=0\) and the covariance matrix has entries \(\Sigma_{ij}=\sigma_X^2c^{|i-j|}\). Unless otherwise specified, we set \(\sigma_X=0.3\) and \(c=0.5\), so adjacent features are positively correlated but the marginal feature scale remains moderate.

The response is generated as \(Y=f(X)+\epsilon\), where \(\epsilon\sim N(0,\sigma_\epsilon^2)\) is independent Gaussian noise with \(\sigma_\epsilon=0.1\). The true regression function is \(f(x)=\sum_{j=1}^{30}x_j^2+\sum_{k=1}^{15}x_{2k-1}x_{2k}\), combining nonlinear main effects with adjacent-pair interactions. This same response function is used in both source and target domains. Therefore, the simulation does not introduce shift by changing the outcome mechanism itself; instead, it introduces shortcut shift through the feature views available to source and target models.

To control the degree of source-target alignment, we vary the number of shared feature dimensions, denoted by \(d_{\mathrm{share}}\), over \(0\), \(10\), \(20\), and \(30\). All models remain defined on the same 30-dimensional input space, but source and target learners are trained with domain-specific feature masks. When \(d_{\mathrm{share}}=0\), the source model uses dimensions \(1\)--\(15\), while the target model uses dimensions \(16\)--\(30\), so the two feature views are disjoint. When \(d_{\mathrm{share}}=10\), the source model uses dimensions \(1\)--\(20\), while the target model uses dimensions \(11\)--\(30\), so the two views share dimensions \(11\)--\(20\). When \(d_{\mathrm{share}}=20\), the source model uses dimensions \(1\)--\(25\), while the target model uses dimensions \(6\)--\(30\), so the two views share dimensions \(6\)--\(25\). When \(d_{\mathrm{share}}=30\), both source and target models use all dimensions \(1\)--\(30\), corresponding to full alignment.

This design creates a gradual transition from severe shortcut shift to complete source-target alignment. At low \(d_{\mathrm{share}}\), source and target models rely on largely different feature subsets, so direct reuse or adaptation of source information may be unreliable. At high \(d_{\mathrm{share}}\), the source and target views become increasingly compatible, and fine-tuning is expected to become more competitive. This makes the simulation useful for evaluating whether \methodname\ can remain effective across different degrees of source-target compatibility.

\subsection{Simulation results}

\begin{figure}[t]
\centering
\includegraphics[width=.75\linewidth]{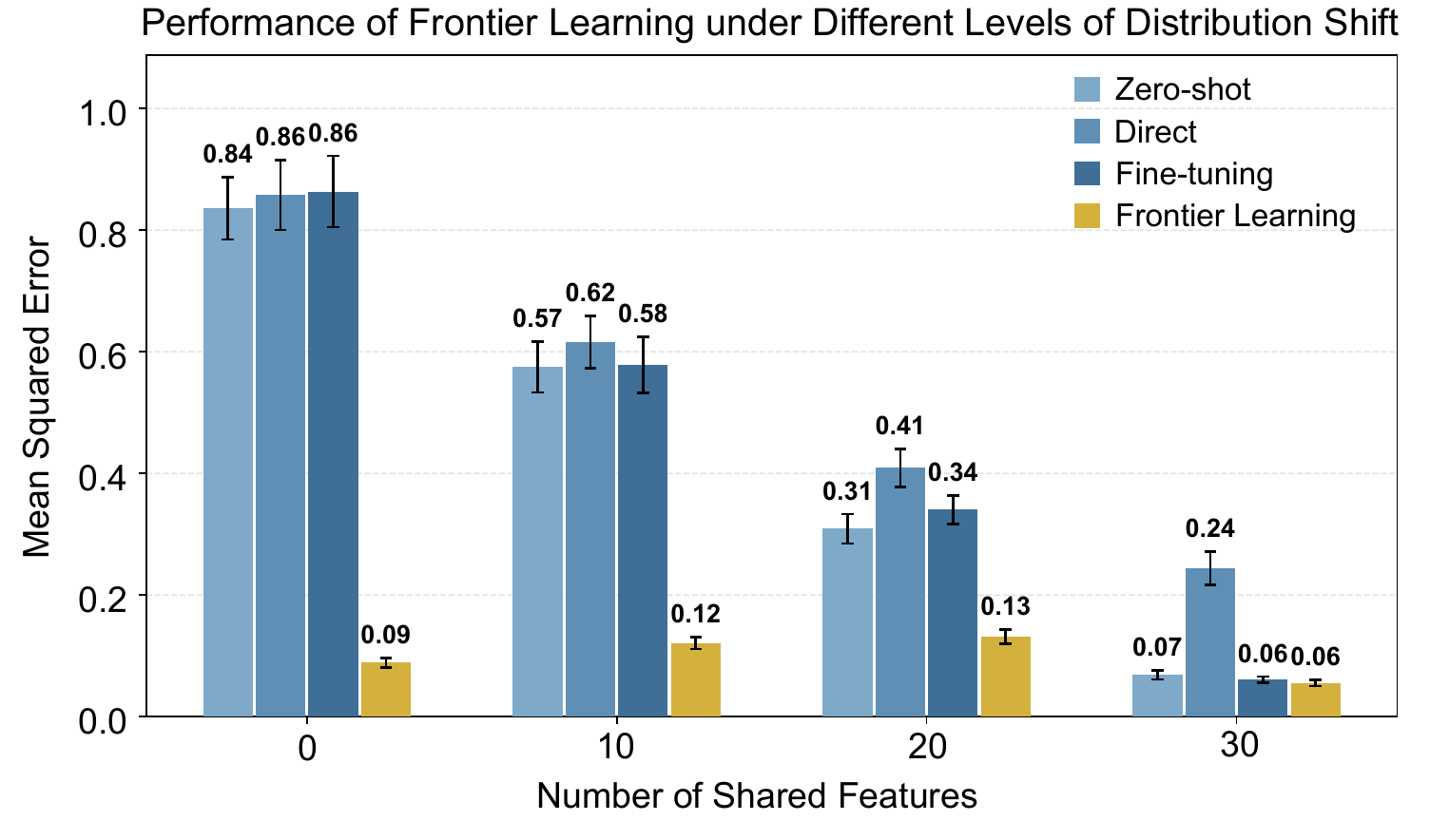}
\caption{Mean squared error (MSE) across varying numbers of shared feature dimensions between source and target models. Lower values indicate better performance. Frontier Learning combines the source zero-shot prediction with hidden representations from fine-tuned and directly trained candidates, achieving the lowest MSE across all shortcut-shift settings.}
\label{fig:simulation}
\end{figure}

Figure~\ref{fig:simulation} summarizes the results. The three individual baselines behave differently as the number of shared dimensions changes. When \(d_{\mathrm{share}}=0\), the source zero-shot model has the lowest MSE among the baselines, with MSE \(0.8359\), compared with \(0.8576\) for direct target training and \(0.8635\) for fine-tuning. When \(d_{\mathrm{share}}=10\), zero-shot remains the strongest individual baseline, with MSE \(0.5746\). When \(d_{\mathrm{share}}=20\), zero-shot reuse again performs best among the baselines, with MSE \(0.3090\). When \(d_{\mathrm{share}}=30\), where source and target feature views are fully aligned, fine-tuning achieves the strongest baseline performance, with MSE \(0.0608\). This variation confirms that no single standard strategy is uniformly best across shortcut-shift settings.

In contrast, \methodname\ achieves the lowest MSE in all four settings. Its MSE values are \(0.0889\), \(0.1212\), \(0.1316\), and \(0.0555\) for \(d_{\mathrm{share}}=0\), \(10\), \(20\), and \(30\), respectively. Relative to the best individual baseline in each setting, this corresponds to MSE reductions of approximately \(89.4\%\), \(78.9\%\), \(57.4\%\), and \(8.8\%\). The improvement is largest when the individual candidate models are weak or capture complementary information, and it becomes smaller when one baseline is already strong, as in the fully aligned setting.

The same pattern is reflected in the secondary metrics. \methodname\ obtains the lowest RMSE and MAE and the highest \(R^2\) in all four settings. For example, when \(d_{\mathrm{share}}=0\), \methodname\ improves \(R^2\) from \(0.4836\) for the best baseline to \(0.9451\). When \(d_{\mathrm{share}}=30\), where the performance gap is smaller, \methodname\ still improves \(R^2\) from \(0.9624\) for fine-tuning to \(0.9657\). These results support the main premise of Frontier Learning: by treating zero-shot reuse, fine-tuning, and direct training as candidate information sources rather than mutually exclusive choices, the frontier learner can adapt to the shift regime and recover a more stable target-domain predictor.

\begin{table}[ht]
\centering
\caption{Full simulation results (mean $\pm$ std over 100 independent repetitions)
across varying numbers of shared feature dimensions ($d_{\text{share}}$).
Bold indicates the best performance in each setting.}
\label{tab:simulation_full}
\resizebox{\textwidth}{!}{%
\begin{tabular}{clcccc}
\toprule
$d_{\text{share}}$ & Method & MSE & RMSE & MAE & $R^2$ \\
\midrule

\multirow{4}{*}{0}
& Zero-shot
& $0.8359 \pm 0.0510$
& $0.9138 \pm 0.0278$
& $0.7035 \pm 0.0180$
& $0.4836 \pm 0.0253$ \\

& Fine-tuning
& $0.8635 \pm 0.0585$
& $0.9287 \pm 0.0314$
& $0.7126 \pm 0.0188$
& $0.4667 \pm 0.0269$ \\

& Direct
& $0.8576 \pm 0.0574$
& $0.9255 \pm 0.0309$
& $0.7123 \pm 0.0185$
& $0.4705 \pm 0.0246$ \\

& Frontier Learning
& $\mathbf{0.0889 \pm 0.0081}$
& $\mathbf{0.2979 \pm 0.0134}$
& $\mathbf{0.2248 \pm 0.0076}$
& $\mathbf{0.9451 \pm 0.0043}$ \\

\midrule

\multirow{4}{*}{10}
& Zero-shot
& $0.5746 \pm 0.0416$
& $0.7575 \pm 0.0272$
& $0.5753 \pm 0.0151$
& $0.6449 \pm 0.0243$ \\

& Fine-tuning
& $0.5784 \pm 0.0464$
& $0.7599 \pm 0.0302$
& $0.5752 \pm 0.0157$
& $0.6426 \pm 0.0266$ \\

& Direct
& $0.6160 \pm 0.0431$
& $0.7844 \pm 0.0272$
& $0.6009 \pm 0.0145$
& $0.6195 \pm 0.0218$ \\

& Frontier Learning
& $\mathbf{0.1212 \pm 0.0098}$
& $\mathbf{0.3478 \pm 0.0140}$
& $\mathbf{0.2662 \pm 0.0092}$
& $\mathbf{0.9251 \pm 0.0055}$ \\

\midrule

\multirow{4}{*}{20}
& Zero-shot
& $0.3090 \pm 0.0242$
& $0.5555 \pm 0.0216$
& $0.4115 \pm 0.0113$
& $0.8089 \pm 0.0153$ \\

& Fine-tuning
& $0.3403 \pm 0.0237$
& $0.5830 \pm 0.0204$
& $0.4348 \pm 0.0125$
& $0.7895 \pm 0.0158$ \\

& Direct
& $0.4090 \pm 0.0313$
& $0.6391 \pm 0.0245$
& $0.4907 \pm 0.0161$
& $0.7474 \pm 0.0159$ \\

& Frontier Learning
& $\mathbf{0.1316 \pm 0.0114}$
& $\mathbf{0.3624 \pm 0.0156}$
& $\mathbf{0.2760 \pm 0.0108}$
& $\mathbf{0.9186 \pm 0.0080}$ \\

\midrule

\multirow{4}{*}{30}
& Zero-shot
& $0.0687 \pm 0.0073$
& $0.2618 \pm 0.0138$
& $0.1977 \pm 0.0082$
& $0.9576 \pm 0.0038$ \\

& Fine-tuning
& $0.0608 \pm 0.0050$
& $0.2464 \pm 0.0100$
& $0.1881 \pm 0.0063$
& $0.9624 \pm 0.0028$ \\

& Direct
& $0.2439 \pm 0.0276$
& $0.4930 \pm 0.0281$
& $0.3818 \pm 0.0202$
& $0.8495 \pm 0.0146$ \\

& Frontier Learning
& $\mathbf{0.0555 \pm 0.0050}$
& $\mathbf{0.2353 \pm 0.0106}$
& $\mathbf{0.1784 \pm 0.0059}$
& $\mathbf{0.9657 \pm 0.0026}$ \\

\bottomrule
\end{tabular}%
}
\end{table}

\section{Real-World Applications}

We evaluate \methodname\ on two real-world distribution-shift settings: visual domain adaptation in DomainNet/VisDA and clinical mortality prediction across ICU domains in MIMIC-IV-Note. These applications test two complementary aspects of the framework. The visual experiment evaluates a mixed-access setting in which white-box models contribute bottleneck representations and a black-box zero-shot model contributes prediction outputs. The clinical NLP experiment evaluates representation-level Frontier Learning across heterogeneous clinical domains.

\subsection{Application to visual domain adaptation in DomainNet/VisDA}

\subsubsection{Setup}

We evaluate \methodname\ on the VisDA-2019 benchmark, which is built on the large-scale DomainNet dataset. DomainNet contains approximately 0.6 million images across 345 object categories and six visually distinct domains: Clipart, Infograph, Painting, Quickdraw, Real, and Sketch \citep{peng:moment}. These domains exhibit substantial shifts in texture, style, abstraction level, and visual complexity, making the benchmark a useful testbed for cross-domain visual prediction.

We consider the supervised domain adaptation setting, where labeled target-domain data are available. We use Painting as the target domain. This choice is motivated by the pairwise zero-shot transfer matrix in Supplementary Table~S1: incoming zero-shot accuracy to Painting varies widely across source domains, ranging from 1.60\% for Quickdraw to 45.68\% for Real. This heterogeneity indicates that different source domains have substantially different compatibility with the target domain, making Painting an informative setting for evaluating whether \methodname\ can integrate heterogeneous candidate models.

All image classifiers use a SHOT-style architecture \citep{liang:we} with a ResNet-50 feature extractor initialized from ImageNet pretrained weights \citep{he:deep}, a 256-dimensional bottleneck layer, and a linear classifier head. The output of the final pooling layer of ResNet-50 is mapped to a bottleneck representation \(r(x)\in\mathbb R^{256}\), which is used as the candidate representation for white-box models.

We consider four candidate models: Direct (Painting), Fine-tuning (Real), Fine-tuning (Clipart), and Zero-shot (Sketch). Direct (Painting) is trained directly on the Painting domain starting from ImageNet initialization. Fine-tuning (Real) and Fine-tuning (Clipart) are initialized from source-domain checkpoints and subsequently fine-tuned on the Painting training split. These two sources are selected because Real and Clipart have the two highest zero-shot transfer accuracies to Painting among all source domains. Zero-shot (Sketch) is used as a black-box zero-shot candidate: the Sketch-trained model is applied directly to Painting without target-domain adaptation, and only its prediction output is used by the frontier learner.

The original Painting training split is divided into training and validation subsets using a 0.9/0.1 split, while the official test split is used for final evaluation. Source models are trained independently on their corresponding source domains using supervised classification. During fine-tuning, we update the last residual block of ResNet-50 together with the bottleneck and classifier layers, while earlier backbone layers remain frozen. Models are trained using SGD with batch size 256. The bottleneck dimension is fixed at 256, label smoothing with parameter 0.1 is applied, and fine-tuning is performed for 30 epochs with learning rate \(10^{-2}\) and weight decay \(10^{-3}\).

For \methodname, we extract 256-dimensional bottleneck representations from the three white-box candidates, Direct (Painting), Fine-tuning (Real), and Fine-tuning (Clipart), and combine them with the prediction output from the black-box Zero-shot (Sketch) candidate. The resulting frontier representation is used to train an \(\ell_2\)-regularized logistic regression classifier on the Painting target-domain training split. During this stage, all candidate feature extractors remain fixed. Hyperparameters are selected using the validation split, and the final model is evaluated once on the held-out target-domain test set. We report top-1 classification accuracy as the primary metric, averaged over ten independent repetitions.

\subsubsection{Results}

\begin{figure}[t]
\centering
\includegraphics[width=.75\linewidth]{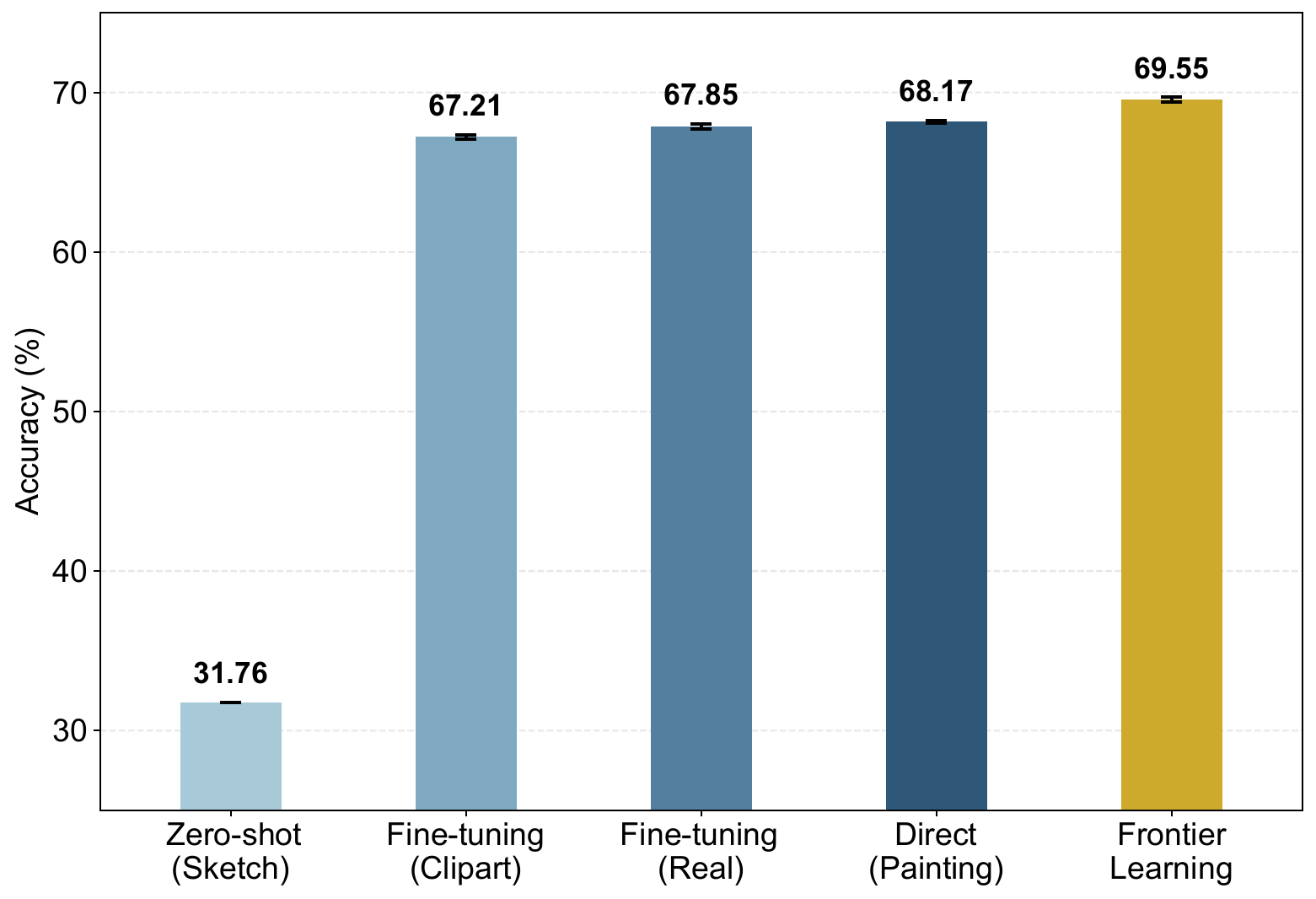}
\caption{Top-1 classification accuracy on the Painting target domain. Frontier Learning combines bottleneck representations from directly trained and fine-tuned white-box candidates together with the prediction output of a black-box zero-shot candidate. The Sketch zero-shot candidate is treated as a fixed black-box model and therefore has zero standard deviation.}
\label{fig:visda}
\end{figure}

Figure~\ref{fig:visda} summarizes the Painting-domain results. The Zero-shot (Sketch) zero-shot candidate performs substantially worse than the other candidates, achieving 31.76\% accuracy. This reflects the considerable visual shift between the Sketch and Painting domains. Among the white-box candidates, Fine-tuning (Real) and Fine-tuning (Clipart) achieve competitive accuracies of 67.85\% and 67.21\%, respectively, but both remain slightly below Direct (Painting), which achieves 68.17\%.

\methodname\ achieves the highest accuracy, 69.55\%, outperforming all individual candidates. Compared with the strongest individual candidate, Direct (Painting), this is an absolute improvement of 1.38 percentage points and a relative classification-error reduction of approximately 4.3\%. The result is notable because the frontier includes a weak zero-shot candidate, yet the target-supervised learner is still able to improve over all individual baselines. This supports the main idea of Frontier Learning: weak source candidates do not need to be selected as final predictors, but their outputs can be included as candidate signals and downweighted when they are not useful for the target task.

\begin{table}[t]
\centering
\caption{Top-1 classification accuracy on the Painting target domain. Frontier Learning combines bottleneck representations from directly trained and fine-tuned white-box candidates together with the prediction output of a black-box zero-shot candidate. The Sketch zero-shot candidate is treated as a fixed black-box model and therefore has zero standard deviation.}
\label{tab:visda}
\begin{tabular}{lc}
\hline
Method & Accuracy \\
\hline
Zero-shot (Sketch) & \(0.3176 \pm 0.0000\) \\
Fine-tuning (Real) & \(0.6785 \pm 0.0016\) \\
Fine-tuning (Clipart) & \(0.6721 \pm 0.0012\) \\
Direct (Painting) & \(0.6817 \pm 0.0009\) \\
\textbf{Frontier Learning} & \(\mathbf{0.6955 \pm 0.0015}\) \\
\hline
\end{tabular}
\end{table}

\subsection{Application to clinical mortality prediction in MIMIC-IV-Note}

\subsubsection{Setup}

We further evaluate \methodname\ on MIMIC-IV-Note, a large-scale collection of deidentified clinical notes linked to the MIMIC-IV electronic health record database \citep{johnson:mimic}. MIMIC-IV-Note contains discharge summaries, radiology reports, nursing notes, and physician notes from patients admitted to Beth Israel Deaconess Medical Center. We use discharge summaries because they summarize the hospitalization trajectory, including diagnoses, procedures, treatments, and outcomes. The prediction task is in-hospital mortality, defined using the hospital expire flag.

Domains are defined according to the patient's first ICU care unit. We focus on five ICU domains: MICU, SICU, CVICU, CCU, and TSICU. Samples are filtered to require a valid discharge summary and a non-missing mortality label. Duplicated notes are removed, and the first ICU care unit is used as the domain identifier. No minimum note length threshold is imposed, and the analysis is not restricted to first ICU stays only. CVICU is selected as the target domain because it has the lowest mean incoming zero-shot AUPRC across source domains in Supplementary Table~S2, indicating that existing source-trained models transfer relatively poorly to CVICU. In the processed data, the CVICU domain contains approximately 11,356 samples with an in-hospital mortality rate of approximately 3.8\%.

We use a hierarchical transformer architecture for long-document prediction \citep{vaswani:attention}. Each discharge summary is divided into non-overlapping chunks of at most 512 tokens, with the maximum number of chunks fixed at 8. Each chunk is encoded using ClinicalBERT \citep{alsentzer:publicly}, which is kept frozen throughout the experiments. The ClinicalBERT embedding for each chunk is projected to a 256-dimensional latent representation, and the sequence of chunk representations is processed by a transformer encoder. An attention-pooling layer then aggregates the contextualized chunk representations into a 256-dimensional note-level representation \(r(x)\), which is used as the candidate representation for Frontier Learning. The projection layer, transformer encoder, attention-pooling layer, and prediction head are trained for the mortality task, while the ClinicalBERT encoder remains frozen.

All models are trained using weighted binary cross-entropy to account for the severe class imbalance in mortality prediction. The positive-class weight is set to the ratio of negative to positive samples. Optimization is performed using AdamW with learning rate \(2\times10^{-5}\), cosine annealing, validation-tuned weight decay, and batch size 1024. Models are trained for up to 100 epochs, with model selection based on validation performance.

We consider four candidate models for the CVICU target domain: Fine-tuning (MICU), Zero-shot (SICU), Zero-shot (CCU), and Direct (CVICU). Direct (CVICU) is trained directly on CVICU target-domain data. Fine-tuning (MICU) is initialized from a MICU source-domain checkpoint and fine-tuned on CVICU. Zero-shot (SICU) and Zero-shot (CCU) are zero-shot candidates, where source-trained models are applied directly to CVICU without adaptation.

Unlike the visual experiment, all candidate models in this clinical NLP experiment are treated as white-box models, including the zero-shot candidates. Here, zero-shot refers to the absence of CVICU adaptation, not to the absence of internal access. Accordingly, \methodname\ extracts the 256-dimensional note-level representation from each candidate and concatenates the four representations into a 1024-dimensional frontier representation. An \(\ell_2\)-regularized logistic frontier learner is then trained on the CVICU training split, while all candidate encoders remain fixed. Model selection is performed using validation AUPRC. We report accuracy, F1 score, AUROC, and AUPRC, with AUPRC treated as the primary metric because in-hospital mortality is rare in this target domain.

\subsubsection{Results}

\begin{figure}[t]
\centering
\includegraphics[width=.75\linewidth]{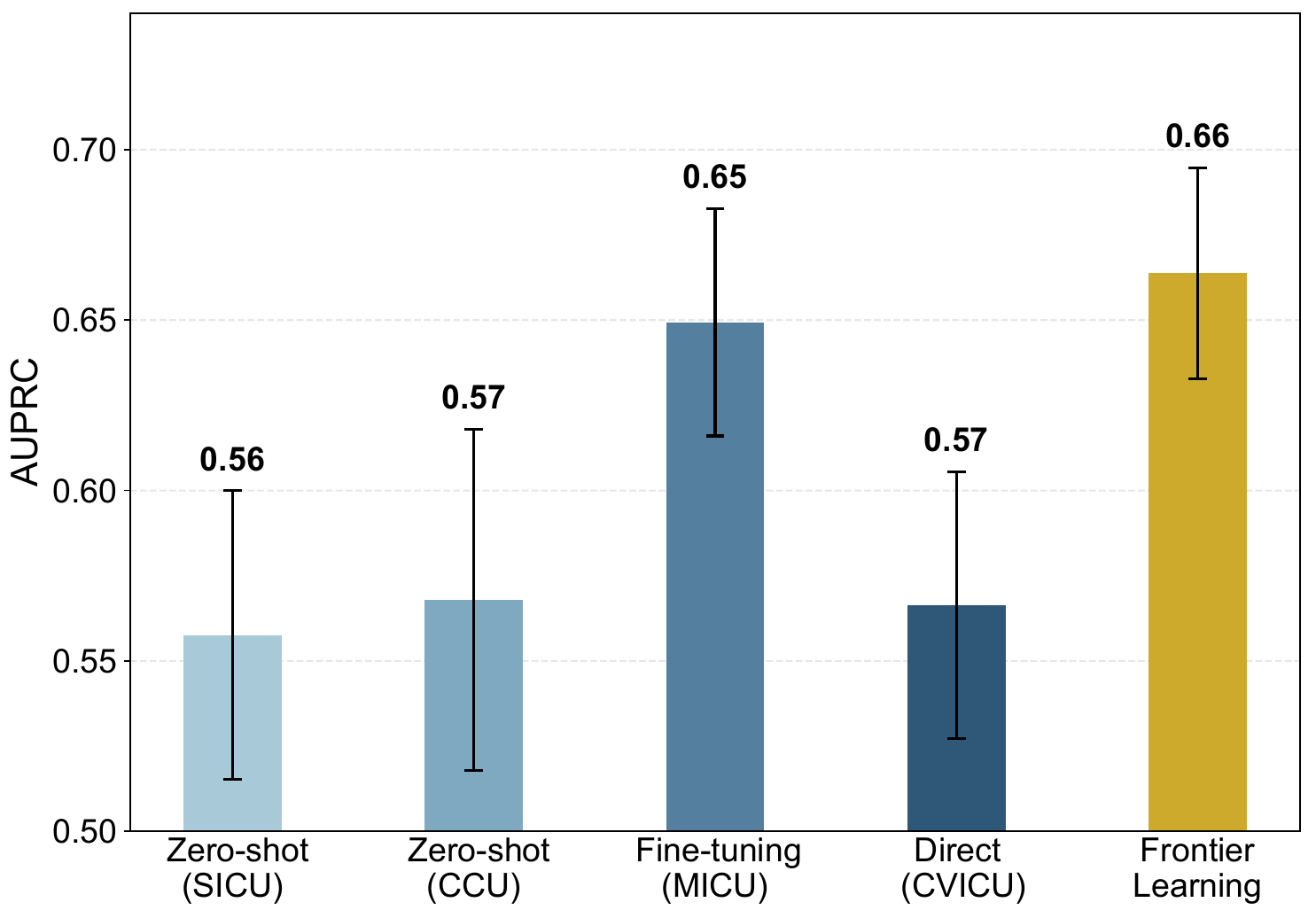}
\caption{AUPRC on the MIMIC-IV-Note CVICU mortality prediction task. Frontier Learning combines note-level representations from fine-tuned, zero-shot, and directly trained ICU-domain candidates and achieves the highest AUPRC among the compared methods.}
\label{fig:mimic}
\end{figure}

Figure~\ref{fig:mimic} summarizes the MIMIC-IV-Note results. Among the individual candidates, Fine-tuning (MICU) achieves the strongest AUPRC, 0.6494, suggesting that MICU provides the most transferable source representation for the CVICU mortality task. The two zero-shot candidates perform worse: Zero-shot (SICU) achieves an AUPRC of 0.5576, and Zero-shot (CCU) achieves an AUPRC of 0.5680. Direct (CVICU) obtains an AUPRC of 0.5664, indicating that direct target-domain training alone is insufficient to match the best transferred candidate under the severe class imbalance in this task.

\methodname\ achieves the highest AUPRC, 0.6638, improving over the best individual candidate by 0.0144 absolute AUPRC, or approximately 2.2\% relative improvement. It also obtains the highest accuracy, 0.9299, and the highest F1 score, 0.4834. The AUROC of \methodname\ is lower than that of Fine-tuning (MICU), indicating that the improvement is not uniform across all ranking metrics. However, because AUPRC is more informative for rare-outcome prediction, the primary result supports the use of Frontier Learning for combining heterogeneous ICU-domain representations in imbalanced clinical prediction.

\begin{table}[t]
\centering
\caption{Classification performance on the MIMIC-IV-Note in-hospital mortality prediction task, reported as mean \(\pm\) standard deviation. AUPRC is the primary evaluation metric due to severe class imbalance. Bold indicates the best performance in each column.}
\label{tab:mimic}
\begin{tabular}{lcccc}
\hline
Method & Accuracy & F1 & AUROC & AUPRC \\
\hline
Zero-shot (SICU) 
& \(0.9235 \pm 0.0063\) 
& \(0.4091 \pm 0.0263\) 
& \(0.9470 \pm 0.0051\) 
& \(0.5576 \pm 0.0423\) \\
Zero-shot (CCU) 
& \(0.8913 \pm 0.0064\) 
& \(0.3718 \pm 0.0228\) 
& \(0.9492 \pm 0.0048\) 
& \(0.5680 \pm 0.0501\) \\
Fine-tuning (MICU) 
& \(0.9086 \pm 0.0194\) 
& \(0.4299 \pm 0.0425\) 
& \(\mathbf{0.9619 \pm 0.0065}\) 
& \(0.6494 \pm 0.0334\) \\
Direct (CVICU) 
& \(0.8845 \pm 0.0108\) 
& \(0.3720 \pm 0.0187\) 
& \(0.9523 \pm 0.0050\) 
& \(0.5664 \pm 0.0391\) \\
\textbf{Frontier Learning} 
& \(\mathbf{0.9299 \pm 0.0220}\) 
& \(\mathbf{0.4834 \pm 0.0612}\) 
& \(0.9452 \pm 0.0187\) 
& \(\mathbf{0.6638 \pm 0.0310}\) \\
\hline
\end{tabular}
\end{table}

\section{Discussion}
In this work, we introduced Frontier Learning, a framework for robust prediction under distribution shift by learning from the complementary information contained in multiple candidate models. Rather than selecting a single source model or adaptation strategy, Frontier Learning combines heterogeneous representations learned under different training regimes and domains into a unified predictor. Our empirical results across simulations, computer vision, and clinical prediction tasks demonstrate that this simple strategy consistently improves predictive performance, particularly when labeled target data are limited and no single candidate model is uniformly optimal.

The success of Frontier Learning suggests that different candidate models often capture distinct aspects of the underlying prediction task. Source-trained models preserve transferable knowledge, fine-tuned models adapt to target-specific characteristics, and directly trained models provide unbiased information from the target domain. By learning over these complementary representations, Frontier Learning is able to exploit useful information that would otherwise be discarded when relying on a single transfer strategy.

Our work also has several limitations. First, the quality of the learned frontier depends on the diversity and usefulness of the candidate model library. If all candidate models produce highly redundant representations, the benefit of aggregation may be limited. Second, the current framework adopts simple feature concatenation followed by a lightweight downstream learner. More sophisticated representation fusion methods, such as attention-based aggregation or adaptive feature selection, may further improve performance while controlling model complexity. Finally, although our experiments demonstrate consistent empirical gains, a theoretical understanding of when representation aggregation improves generalization under distribution shift remains an important direction for future work.

The increasing availability of pretrained foundation models further broadens the applicability of Frontier Learning. As modern machine learning systems increasingly consist of collections of heterogeneous black-box and white-box models, learning over multiple pretrained models rather than selecting a single one may become a practical paradigm for robust adaptation under distribution shift. We hope this work motivates further investigation into representation-level aggregation as a general framework for transfer learning.

\section{Software}
The implementation of 
frontier learning
is available at \url{https://github.com/batmen-lab/frontier_learning}.

\section{Supplementary Material}

\paragraph{VisDA-2019.}
The VisDA-2019 benchmark contains images from six visual domains (clipart, infograph, painting, quickdraw, real, and sketch) covering 345 object categories. In our experiments, Painting is treated as the target domain, while models trained on other domains serve as candidate source models. The dataset is available at \url{http://ai.bu.edu/M3SDA/}.

\paragraph{MIMIC-IV.}
MIMIC-IV is a publicly available electronic health record database. We use the discharge notes associated with ICU admissions as the input text and define different ICU units as distinct domains for transfer learning experiments. The dataset is available at \url{https://physionet.org/content/mimiciv/}.

\section*{SUPPLEMENT}
\renewcommand{\thetheorem}{S\arabic{theorem}}
\renewcommand{\thelemma}{S\arabic{lemma}}
\renewcommand{\theremark}{S\arabic{remark}}
\renewcommand{\theequation}{S\arabic{equation}}
\renewcommand{\thesection}{S\arabic{section}}
\renewcommand{\thetable}{S\arabic{table}}
\def\theHsection{S\arabic{section}}
\def\theHtheorem{ST\arabic{section}.\arabic{theorem}}
\def\theHlemma{ST\arabic{section}.\arabic{theorem}}
\def\theHremark{SR\arabic{section}.\arabic{remark}}
\def\theHequation{SE\arabic{equation}}
\def\theHtable{STB\arabic{table}}
\setcounter{theorem}{0}
\setcounter{equation}{0}
\setcounter{section}{0}
\setcounter{table}{0}

\section{Additional Experiment Details}
\subsection*{VisDA: Zero-Shot Transfer Matrix and Experimental Design}

Table~\ref{tab:visda_zeroshot_matrix} reports the pairwise zero-shot top-1 
accuracy (\%) across all six DomainNet domains, where entry $(s, t)$ denotes 
the accuracy achieved by a model trained on domain $s$ and applied directly 
to domain $t$ without any adaptation.

\begin{table}[ht]
\centering
\caption{Pairwise zero-shot transfer accuracy (\%) across DomainNet domains.
Rows denote source domains and columns denote target domains.
Diagonal entries are left blank.}
\label{tab:visda_zeroshot_matrix}

\begin{tabular}{lcccccc}
\toprule
 & Clipart & Infograph & Painting & Quickdraw & Real & Sketch \\
\midrule
Clipart    & ---     & 14.4657 & 33.9726 &  9.6406 & 50.3904 & 37.7227 \\
Infograph  & 26.0024 & ---     & 27.9486 &  1.4512 & 42.6446 & 22.4390 \\
Painting   & 36.9988 & 15.1793 & ---     &  2.2821 & 55.2801 & 33.2189 \\
Quickdraw  &  8.4042 &  0.9805 &  1.5988 & ---     &  4.2737 &  8.2742 \\
Real       & 45.7608 & 18.1147 & 45.6841 &  4.8019 & ---     & 33.7549 \\
Sketch     & 44.5187 & 11.9461 & 31.7622 & 10.2357 & 44.2878 & ---     \\
\midrule
Column mean & 32.3370 & 12.1373 & 28.1933 & 5.6823 & 39.3753 & 27.0817 \\
\bottomrule
\end{tabular}

\end{table}

\paragraph{Choice of target domain.}
We select Painting as the target domain because it exhibits the greatest 
heterogeneity in incoming zero-shot accuracy across source domains, 
ranging from $1.60\%$ (Quickdraw) to $45.68\%$ (Real). 
This wide spread indicates that different source domains vary substantially 
in their compatibility with Painting, making it an informative testbed for 
evaluating whether Frontier Learning can effectively integrate heterogeneous 
source information.

\paragraph{Choice of fine-tuning sources.}
Among all source domains, Fine-tuning (Real) and 
Fine-tuning (Clipart) achieve the two highest zero-shot 
accuracies ($45.68\%$ and $33.97\%$, respectively), indicating that these 
two domains are the most compatible with Painting in terms of visual 
characteristics. We therefore select Real and Clipart as fine-tuning 
sources, as initializing from the most compatible source checkpoints 
is expected to yield the strongest adapted models.

\paragraph{Choice of zero-shot source.}
We include Zero-shot (Sketch) as the black-box 
zero-shot baseline. Sketch achieves a moderate zero-shot accuracy of 
$31.76\%$ to Painting, offering complementary stylistic information 
distinct from the photorealistic Real and the stylized Clipart domains. 
Infograph and Quickdraw are excluded due to their substantially lower 
zero-shot performance ($27.95\%$ and $1.60\%$, respectively), suggesting 
minimal transferable information to the Painting domain.

\subsection*{MIMIC-IV-Note: Zero-Shot Transfer Matrix and Experimental Design}

Table~\ref{tab:zeroshot_matrix} reports the pairwise zero-shot AUPRC across all 
five ICU domains, where entry $(s, t)$ denotes the AUPRC achieved by a model 
trained on domain $s$ and applied directly to domain $t$ without any adaptation.

\begin{table}[t]

\centering

\caption{Zero-shot performance (AUPRC) across ICU domains.}

\label{tab:zeroshot_matrix}

\begin{tabular}{lccccc}
\toprule
 & MICU & CVICU & SICU & CCU & TSICU \\
\midrule
MICU   & ---    & 0.5930 & 0.6970 & 0.7791 & 0.6025 \\
CVICU  & 0.6276 & ---    & 0.6291 & 0.6983 & 0.5605 \\
SICU   & 0.6932 & 0.5631 & ---    & 0.7369 & 0.6405 \\
CCU    & 0.7063 & 0.5802 & 0.6760 & ---    & 0.5774 \\
TSICU  & 0.6674 & 0.5656 & 0.7004 & 0.7124 & ---    \\
\midrule
Column mean & 0.6736 & 0.5755 & 0.6759 & 0.7317 & 0.5952 \\
\bottomrule
\end{tabular}

\end{table}

\paragraph{Choice of target domain.}
We select CVICU as the target domain because it receives the lowest mean 
zero-shot AUPRC across all source domains ($0.576$), indicating that 
existing source-trained models transfer poorly to CVICU. 
This makes CVICU the most challenging and informative target for 
evaluating transfer learning robustness.

\paragraph{Choice of fine-tuning source.}
Among all source domains, Fine-tuning (MICU) achieves the 
highest zero-shot AUPRC ($0.593$), suggesting that MICU is the most 
clinically compatible source for the CVICU target. 
We therefore select MICU as the single fine-tuning source, as initializing 
from the most compatible source checkpoint is expected to yield the strongest 
adapted model.

\paragraph{Choice of zero-shot sources.}
We include Zero-shot (SICU) and 
Zero-shot (CCU) as zero-shot baselines to provide 
complementary source-domain representations within the Frontier Learning 
ensemble. TSICU is excluded to keep the ensemble size manageable, as its 
zero-shot AUPRC to CVICU ($0.566$) is comparable to that of SICU ($0.563$) 
and offers limited additional diversity. Together, the four models—Fine-tuning (MICU), Zero-shot (SICU), Zero-shot (CCU), and Direct (CVICU)—cover a range of training regimes and source–target compatibility levels, providing a diverse candidate library for Frontier Learning.

\bibliographystyle{apalike} 
\bibliography{refs}

\end{document}